\documentclass[sigconf]{acmart}
\usepackage{tikz}
\usetikzlibrary{arrows.meta,positioning,calc,fit,backgrounds,shapes.geometric}
\usepackage{balance}

\copyrightyear{2026}
\acmYear{2026}
\setcopyright{cc}
\setcctype{by}
\acmConference[UbiComp Companion '26]{Companion of the 2026 ACM International Joint Conference on Pervasive and Ubiquitous Computing}{October 11--15, 2026}{Shanghai, China}
\acmBooktitle{Companion of the 2026 ACM International Joint Conference on Pervasive and Ubiquitous Computing (UbiComp Companion '26), October 11--15, 2026, Shanghai, China}
\acmDOI{10.1145/3798063.3837307}
\acmISBN{979-8-4007-2533-3/2026/10}

\usepackage{fancyhdr}   

\fancypagestyle{firstpage}{     
  \fancyhf{}
  \chead{\textcolor{gray}{This article has been accepted for publication in the Companion of the 2026 ACM International \\ Joint Conference on Pervasive and Ubiquitous Computing.}}
  \fancyfoot[C]{\small{\textcolor{gray}{~\copyright~ Personal use of this material is permitted.  Permission from ACM must be obtained for all other uses, in any current or future media, including reprinting/republishing this material for advertising or promotional purposes, creating new collective works, for resale or redistribution to servers or lists, or reuse of any copyrighted component of this work in other works.}}}

}

\begin{document}

\title{You Don't Need To Train: Agentic Heuristic Learning Studio for Executable Human Activity Recognition}

\author{Siyu Yuan}
\affiliation{%
  \institution{RPTU University Kaiserslautern-Landau}
  \city{Kaiserslautern}
  \state{Rheinland Pfalz}
  \country{Germany}}

\author{He Zhang}
\affiliation{%
  \institution{Northwestern Polytechnical University}
  \city{Xi'an}
  \state{Shaanxi}
  \country{China}}

\author{Sizhen Bian}
\affiliation{%
  \institution{Northwestern Polytechnical University}
  \city{Xi'an}
  \state{Shaanxi}
  \country{China}}

\author{Bin Guo}
\affiliation{%
  \institution{Northwestern Polytechnical University}
  \city{Xi'an}
  \state{Shaanxi}
  \country{China}}

\renewcommand{\shortauthors}{Siyu Yuan, He Zhang, Sizhen Bian, and Bin Guo}

\begin{abstract}
Human activity recognition (HAR) is usually framed as gradient-based training of neural networks. Agentic Heuristic Learning (AHL) Studio explores a complementary view inspired by human cognitive learning: people learn activities by remembering examples, forming rules, and repairing mistakes, not by backpropagating. This proposed tool implements AHL for HAR: a learning-time agent reasons over sensor protocols, proposes executable heuristic policies, records repair traces, and exports an LLM-free policy for edge deployment. We focus on the HAR benchmark family and provide an end-to-end workflow from dataset observation to edge-oriented export. On eleven HAR datasets evaluated so far, AHL policies reach strong executable-policy performance while remaining inspectable, editable, and replayable \footnote{https://github.com/zhaxidele/ahl-ts-studio}.
\end{abstract}



\maketitle

\begin{figure*}[t!]
\centering
\includegraphics[width=\textwidth, height=5.8cm]{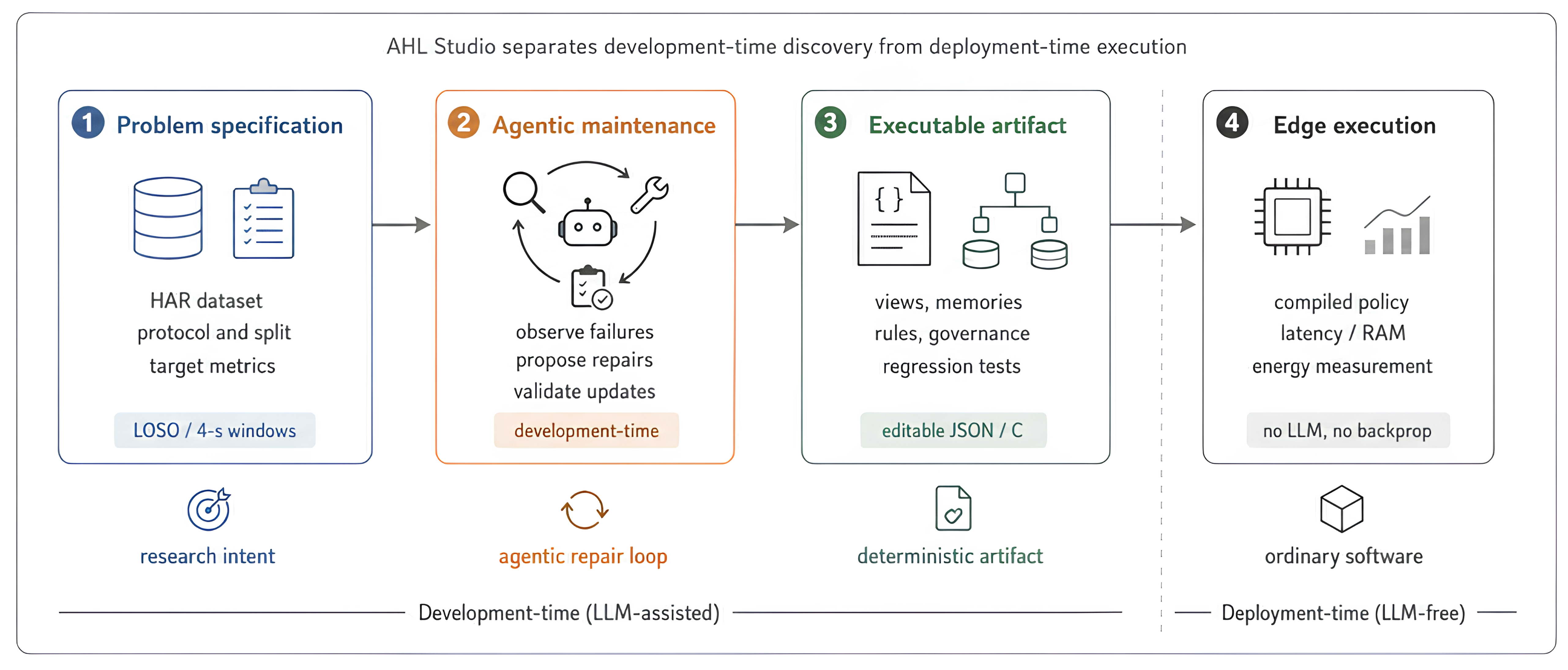}
\caption{AHL Studio provides an end-to-end HAR workflow. 
}
\label{fig:workflow}
\end{figure*}

\section{Introduction}
\thispagestyle{firstpage} 
HAR is a natural domain for studying learning objects other than neural weights, in the spirit of heuristic learning and reflective language agents~\cite{lenat1983role, weng2026learning_beyond_gradients,shinn2023reflexion,nanashima2021theory,gigerenzer2011heuristic,reiss2002learning,hwang2010heuristic}. A person learns ``walking'', ``sitting'', or ``lifting'' by comparing new motion to remembered examples, forming reusable rules, and correcting mistakes after feedback \cite{celemin2019reinforcement,vanlehn1999rule, vanlehn1998analogy,hannafin1993learning}. This kind of learning uses memory, abstraction, and repair; it does not require backpropagation \cite{werbos1988backpropagation}. AHL Studio turns this intuition into a tool: the learning-time agent reads a dataset description, reasons about sensors and protocols, proposes executable heuristics, and records why each repair is expected to help.
This view is useful for wearable sensing because HAR systems are often deployed under tight constraints\cite{zhang2026synthesizing}. A model may need to run on a microcontroller, adapt to a new body position, expose why an activity was predicted, or be revised after a researcher notices a failure case\cite{liu2026cool, ek2025comparing, konwar2025revisiting}. Deep neural networks are powerful, but the learned object is usually an opaque parameter vector\cite{zhang2026synthesizing,yuan2024exploring, saha2025feature, ek2025comparing,konwar2025revisiting,moosmann2024ultra, saha2025feature, dhekane2025transfer}. AHL Studio instead treats the learned object as a maintained executable system: a policy plus memory, rules, regression checks, repair traces, and compression records.
The important boundary is that the agent is not deployed. The final artifact is a compact policy consisting of feature views, temporal voting, subject calibration, prototype memories, tree rules, regression tests, and compression records. It can be inspected, edited, replayed, and exported as ordinary software. This distinguishes AHL Studio from both hand-written rules and black-box neural training: the system learns through agent-assisted maintenance, but inference remains LLM-free.
The submission makes four contributions. First, it frames HAR model development as agentic heuristic learning, where memory, rules, feedback, and repair are explicit parts of the learned system. Second, it presents AHL Studio, an end-to-end tool that connects dataset and paper ingestion, grammar-guided policy discovery, editable policy replay, and STM32-oriented deployment export. Third, it introduces transparent learning-cost accounting by logging LLM prompts, responses, token use, fallback states, and deployment-time separation from LLM inference. Fourth, it evaluates executable AHL policies across HAR benchmarks and compares them with representative neural baselines.

\section{Related Work}
Wearable HAR has been dominated by neural architectures that learn temporal
representations from windowed sensor streams\cite{chen2021deep,bonazzi2024retina,guo2026calibration, kulsoom2022review,bian2026foundation, gupta2022human}. DeepConvLSTM is a widely used
convolutional-recurrent baseline \cite{ordonez2016deep}, TinyHAR improves efficiency through compact
temporal modeling\cite{zhou2022tinyhar}, and TinierHAR further reduces parameters and MACs while
maintaining average macro F1 across a broad HAR benchmark suite~\cite{bian2025tinierhar}.
These methods are strong deployment-oriented neural baselines, but their
learned object is still a parameter vector that is difficult to inspect or
repair directly.
Classical HAR already used hand-crafted descriptors, nearest prototypes,
threshold rules, and decision trees \cite{sargano2017comprehensive,shaban2025comparative}, but those systems were usually manually
designed and did not maintain explicit repair traces. Recent progress in large
language models and reflective agents makes it possible to revisit heuristic
systems as a learning object \cite{weng2026learning_beyond_gradients,shinn2023reflexion}. In this view, an agent can read dataset
descriptions and related papers, reason about sensing protocols, propose repairs, and
record why a candidate policy should help. The final deployed artifact,
however, remains ordinary executable software rather than an LLM or a neural
runtime.
Our proposal is therefore organized around the most explored HAR
benchmark family. This choice makes the tool
concrete: rather than presenting AHL Studio as a generic machine-learning
interface, the challenge version focuses on the real protocol and deployment
issues faced by wearable HAR researchers.

\section{End-to-End Tool}
Figure~\ref{fig:workflow} summarizes the full path from dataset to edge deployment. A researcher specifies a HAR dataset, protocol, target metrics, and related SOTA papers. The app extracts the dataset and paper notes, recommends grammar primitives, runs multi-round LLM-assisted repair, displays maintenance traces, exposes the final policy as editable JSON, and exports an MCU-oriented deployment package.

The user workflow has five stages. First, the user provides the HAR dataset path, protocol, target metric, and optional dataset notes. Second, the user uploads related SOTA papers; the tool extracts relevant protocol assumptions, model baselines, and target metrics. Third, the user selects or accepts recommended grammar primitives. Fourth, the learning-time agent performs multi-round proposal and repair. Finally, the selected executable policy can be edited, replayed, and exported to an embedded project folder.

\subsection{Heuristic Grammar}
AHL Studio uses a bounded grammar of HAR primitives. Temporal primitives include sequence voting, smoothing, hysteresis, and minimum-duration filters. Motion primitives include signal magnitude area, energy, frequency-domain summaries, and compact statistical descriptors. Semantic primitives include subject calibration, body-position assumptions, and tree-rule ensembles. Memory/prototype substrates include class centroids, medoids, compact prototypes, and metric memories. Each primitive is annotated in the interface with expected runtime cost, memory load, and interpretability.
The grammar is not meant to hide domain knowledge. Instead, it makes domain knowledge explicit and executable. The LLM does not write arbitrary code; it proposes repairs within the selected primitive set. This design keeps the learning process auditable and prevents deployment from depending on an LLM runtime.

\subsection{Agentic Maintenance Loop}
Each learning round contains observation, diagnosis, repair, validation, and compression. The prompt includes the dataset summary, SOTA target, selected grammar, previous feedback, and active repair memory. The response is expected to include a failure hypothesis, reasoning, candidate policy deltas, regression risks, and a compression plan. The tool records the exact prompt, raw response, parsed policy proposal, validation event, and promotion decision.
This loop is designed to make the ``thinking'' part visible. Instead of showing only a final score, AHL Studio shows why a proposal was made: for example, temporal voting may be proposed when predictions flicker between adjacent activities, subject calibration may be proposed when held-out-user performance drops, and compact prototypes may be proposed when memory retrieval is helpful but the deployment footprint must remain small.

\subsection{Interface Components}
\begin{figure*}[t]
\centering
\includegraphics[width=\textwidth, height=5.7cm]{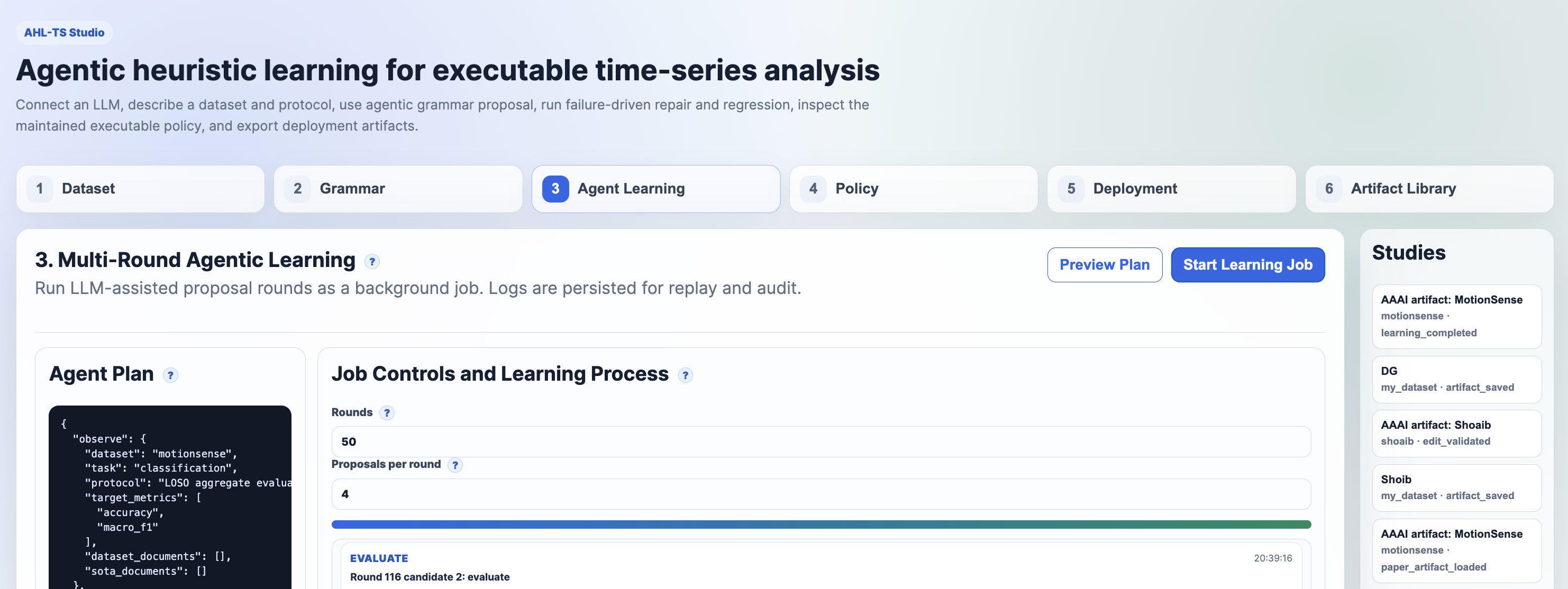}
\caption{
AHL Studio interface for time-series analysis (HAR).
The tool supports dataset and protocol specification, agentic grammar proposal,
agentic failure-driven learning, executable policy inspection, and deployment
artifact export.
}
\label{fig:ahl-studio-interface}
\end{figure*}

Figure~\ref{fig:ahl-studio-interface} shows the prototype AHL Studio interface
used to organize researcher-facing time-series workflows, from dataset
specification to executable policy export. The goal is to make the researcher complete the same steps that would otherwise be scattered across scripts, notebooks, API calls, and embedded projects. Every page writes an artifact: extracted document summaries, selected grammar primitives, LLM prompt-response records, candidate logs, final policy JSON, and deployment package metadata.
The policy editor is particularly important. A final policy is not just a label such as ``random forest'' or ``prototype classifier''. It is a serialized artifact that exposes parameters such as feature views, prototype count, vote window, threshold, tree depth, compression mode, and regression-gate settings. This is the part of the tool where the user can apply human repair after the agent has proposed a policy.


\section{Learning Cost and Energy Accounting}
AHL Studio treats learning cost as a first-class output. Each run stores the model provider, prompt text, raw response, token usage, and whether a response came from the external LLM or a fallback mode. 
Table~\ref{tab:cost} reports a recent audit record. 
Because provider-side energy varies by model and serving hardware, we report both raw token statistics and a configurable token-energy proxy $E_{\mathrm{learn}}=N_{\mathrm{ktok}}e_{\mathrm{ktok}}$, where $e_{\mathrm{ktok}}$ is energy per thousand tokens. The key deployment point is independent of the coefficient: after learning, the exported HAR policy consumes zero LLM tokens.
The token count is not presented as a substitute for energy measurement. Instead, it is an audit trail that makes learning-time resource use visible. If a local LLM is used, the same accounting can be paired with measured GPU or CPU energy. If a commercial API is used, the record provides a model-independent usage summary that can be combined with provider pricing or estimated energy per token. In both cases, AHL Studio separates learning-time cost from inference-time cost, which is crucial for wearable deployment.


\begin{table}[t]
\caption{Learning-cost record from the AHL Studio prototype.}
\label{tab:cost}
\scriptsize
\begin{tabular}{ll}
\toprule
Item & Logged value \\
\midrule
External LLM rounds & 100/100 responded \\
Fallback rounds & 0 \\
Document/API requests snapshot & 8 requests \\
Input tokens, cache hit & 17,024 \\
Input tokens, cache miss & 29,903 \\
Output tokens & 8,378 \\
Total tokens & 55,305 \\
Token-energy accounting & $55.305 e_{\mathrm{ktok}}$ \\
Deployment LLM tokens & 0 \\
\bottomrule
\end{tabular}
\end{table}

\section{Editable Policy Example}
One advantage of executable policies is that they can be locally repaired. For example, if a MotionSense policy shows label flicker between walking and jogging, the agent may propose increasing the temporal vote window. The researcher can inspect the proposal, accept it, or edit the value manually. The resulting change is small and replayable:
\begin{verbatim}
before: { "feature_view": "har_stats",
          "memory": {"type": "prototype", "k": 5},
          "rules": {"vote_window": 5} }
after:  { "feature_view": "har_stats",
          "memory": {"type": "prototype", "k": 7},
          "rules": {"vote_window": 9} }
\end{verbatim}
This example illustrates the intended repair granularity. The user does not retrain a neural network, regenerate all candidates, or rewrite an embedded project. Instead, the user edits an executable field, reruns regression checks, and verifies that protected cases remain valid. This is where the human-learning analogy becomes practical: the system retains memories and rules, receives feedback, and updates a localized part of the behavior.

\begin{table}[t]
\caption{HAR results (F1), neural-baseline comparison, and STM32 deployment cost.}
\label{tab:results}
\scriptsize
\setlength{\tabcolsep}{1.4pt}
\begin{tabular}{p{1.05cm}p{0.82cm}p{0.48cm}p{0.48cm}p{0.50cm}p{0.72cm}p{0.46cm}p{0.48cm}p{0.48cm}p{1.65cm}}
\toprule
Dataset & Protocol & AHL & Tinier HAR & Tiny HAR & DeepCon vLSTM & $\Delta$ & Lat. ms & RAM kB &  Policy \\
\midrule
MotionSense & LOSO & \textbf{96.04} & 92.1 & 91.4 & 91.6 & +3.94 & 5.1 & 1.1  & tree rules, selected HAR views \\
RecGym & wrist & 91.52 & 92.0 & 91.8 & \textbf{92.4} & -0.88 & 7.8 & 1.5   & compact HAR views + tree rules \\
Shoaib & LOSO & \textbf{100.00} & 98.9 & 99.2 & 99.9 & +0.10 & 6.2 & 2.4   & spectral metric memory + vote \\
UCI HAR & official & 92.78 & \textbf{95.5} & 95.1 & 95.1 & -2.72 & 0.8 & 0.9   & official + invariant views \\
DSADS & LOSO & \textbf{96.97} & 86.6 & 85.8 & 84.8 & +10.37 & 5.7 & 1.2   & HAR36 tree rules, 64 trees, depth 8 \\
HAPT & official & 82.80 & 82.6 & 82.2 & \textbf{85.4} & -2.60 & 6.9 & 1.3 &  rich tree rules, depth 12 \\
MHEALTH & LOSO & \textbf{96.60} & 92.3 & 90.2 & 90.1 & +4.30 & 0.6 & 1.0   & centroid memory, HAR36 view \\
PAMAP2 & LOSO & \textbf{88.80} & 76.9 & 76.0 & 75.9 & +11.90 & 0.9 & 1.4  & centroid memory, spectral view \\
SHO & strat. & 98.92 & 94.6 & \textbf{99.2} & 94.8 & -0.28 & 8.5 & 1.5  & rich tree rules, depth 16 \\
USCHAD & LOSO & \textbf{84.50} & 73.5 & 74.8 & 75.4 & +9.10 & 16.8 & 2.5   & spectral tree rules, 128 trees \\
WISDM & strat. & \textbf{89.31} & 81.3 & 82.8 & 82.9 & +6.41 & 8.3 & 1.5  & rich tree rules, depth 16 \\
\bottomrule
\end{tabular}
\end{table}

\section{HAR Results}
Table~\ref{tab:results} compares AHL Studio with neural HAR baselines. AHL Studio is not claiming to
dominate neural HAR models on every dataset. Instead, it shows that an
agent-maintained executable policy can often be strong enough to be useful
while offering properties that neural baselines usually do not expose: a
readable policy, repair traces, explicit regression checks, and direct
deployment export. Across the ten datasets with reported neural baselines in
Table~\ref{tab:results}, AHL-TS outperforms the strongest listed neural
baseline on six datasets, remains within one macro-F1 point on RecGym and SHO,
and underperforms on UCI HAR and HAPT. The selected policies are also not
uniform. MHEALTH and PAMAP2 select centroid-memory policies, while DSADS, HAPT,
SHO, USCHAD, and WISDM require tree-rule partitions. This diversity is useful
evidence that AHL Studio is choosing among heuristic families according to
dataset behavior, rather than applying one AutoML template.

\section{Edge Deployment}
The deployment page exports the final policy, memory/prototype tables, and a small benchmark harness (Table~\ref{tab:results}). For STM32-class boards, the exported harness measures cycles, latency, working RAM, and table size. The current implementation targets STM32N657X0, but the policy itself is ordinary C-style logic: feature extraction, voting, prototype lookup, or tree-rule evaluation. Because no LLM is called at inference, the runtime cost is determined only by the selected primitive family and the size of retained memories.

\section{Maintenance Trace and Replay}
AHL Studio records the learning process as a sequence of replayable maintenance
events rather than only reporting the final score. Each event stores the
observed failure summary, the grammar primitives that were active, the agent's
repair rationale, the candidate policy delta, validation metrics, regression
status, and compression action. This trace is central to the AHL formulation:
the system does not merely search for a high-scoring policy, but maintains an
executable artifact whose future updates are constrained by accumulated repair
memory and regression tests.

Table~\ref{tab:trace-example} gives a compact example from the MotionSense
study. The initial policy used a shallow tree-rule family over HAR views. The
agent then diagnosed cross-subject confusion between dynamic activities and
promoted repairs that changed the feature view and governance window. Later
rounds rejected larger tree ensembles when the gain was too small relative to
the executable-complexity penalty. The final promoted artifact was therefore a
compact tree-rule policy rather than the largest candidate in the grammar.

\begin{table}[t]
\caption{Representative maintenance trace for MotionSense. 
}
\label{tab:trace-example}
\scriptsize
\begin{tabular}{p{0.35cm}p{2.55cm}p{2.25cm}p{0.75cm}}
\toprule
Step & Failure mode & Accepted repair & Macro F1 \\
\midrule
0 & dynamic classes confused across users & initialize HAR statistical view with tree rules & 0.9371 \\
3 & short-window jitter between adjacent activities & add temporal governance and selected HAR views & 0.9528 \\
5 & residual subject-specific boundary errors & increase tree depth under regression gate & 0.9604 \\
7 & larger ensemble gives negligible gain & compress candidate history; retain compact policy & 0.9604 \\
\bottomrule
\end{tabular}
\end{table}

Replayability is implemented by separating three artifacts. First, the
\emph{discovery record} contains prompt-response logs and the natural-language
repair hypotheses. Second, the \emph{candidate log} contains deterministic
policy configurations and their validation outcomes. Third, the
\emph{deployment artifact} contains only the promoted executable policy and the
tables needed for inference. This separation allows reviewers or users to audit
where the agent contributed reasoning, while rerunning the reported benchmark
without invoking the LLM.

\begin{table}[t]
\caption{Replay artifacts produced by AHL Studio.}
\label{tab:replay-artifacts}
\scriptsize
\begin{tabular}{ll}
\toprule
Artifact & Stored content \\
\midrule
Discovery record & dataset summary, SOTA notes, prompt, LLM response \\
Candidate log & grammar choice, policy delta, score, regression status \\
Repair memory & rejected repairs, accepted fixes, failure summaries \\
Regression suite & protected cases and minimum accepted scores \\
Deployment artifact & executable policy JSON, memory tables, MCU harness \\
\bottomrule
\end{tabular}
\end{table}

This structure also supports human repair. A user can edit a localized field in
the final policy, rerun the regression suite, and compare the edited artifact
with the promoted version. Thus, AHL Studio exposes not only a model output, but
also the maintenance state that explains how the executable HAR policy was
constructed, repaired, and compressed.

\section{Limitations and Next Steps}
The current tool does not discover all possible primitives from scratch; it reasons over a bounded HAR grammar. The energy accounting currently separates measured MCU deployment cost from token-normalized learning cost. Another limitation is that LLM-assisted learning can be slow when the provider is overloaded or when prompts are too long; the tool now records provider responses and fallback states to make this visible.

Next steps are to add direct hardware-in-the-loop energy measurement, and run user studies measuring whether AHL Studio reduces the time required to debug and deploy HAR systems. We also plan to add local LLM support for privacy-sensitive wearable datasets, stronger policy compression, and richer visual explanations of activity-level and subject-level failure modes.

\section{Conclusion}
AHL Studio reframes HAR model development as agent-assisted heuristic learning rather than only gradient training. It connects human-like memory and repair, end-to-end dataset-to-edge tooling, transparent token/energy accounting, and executable policies that remain editable after learning. The result is a practical research tool for exploring HAR systems that are not only accurate, but also inspectable, replayable, and deployable. 

\begin{acks}
This work is supported by “the Fundamental Research Funds for the Central Universities.
\end{acks}

\balance
\bibliographystyle{ACM-Reference-Format}
\bibliography{sample-base}

\end{document}